\newif\ifisfinal\isfinaltrue
\documentclass[11pt]{article}

\usepackage{preamble}

\begin{document}
\zavenbegindoc%

\begin{abstract}
  The Achilles heel of \Acfp{llm} is hallucination, which has drastic
  consequences for the clinical domain.  This is particularly important with
  regards to automatically generating discharge summaries (a lengthy medical
  document that summarizes a hospital in-patient visit). Automatically
  generating these summaries would free physicians to care for patients and
  reduce documentation burden.
  The goal of this work is to discover new methods that combine language-based
  graphs and deep learning models to address provenance of content and
  trustworthiness in automatic summarization.
  Our method shows impressive reliability results on the publicly available
  \mimic\ corpus and clinical notes written by physicians at \uihealthname.  We
  provide our method, generated \ds\ output examples, source code and trained
  models.
\end{abstract}
\acresetall 

\zfadd[][0.85\columnwidth]{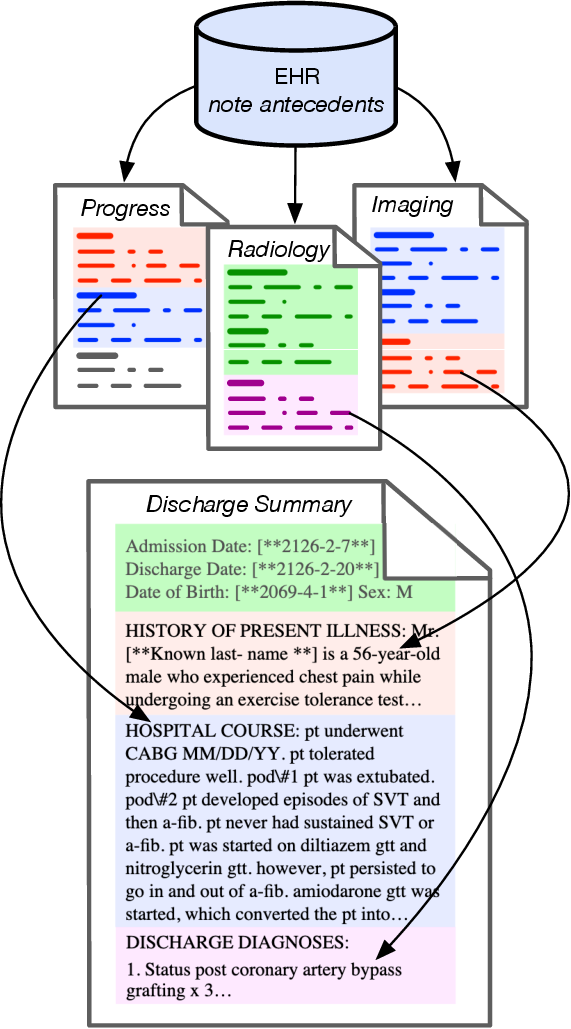}[Discharge Summary Generation]{Clinical
  notes are used as source text to automatically write a hospital discharge
  summary. Topic specific document is colored by section.}

\zssec{Introduction}

Automatic summarization is the task of using machines to summarize natural
language
text~\cite{maynezFaithfulnessFactualityAbstractive2020,ranjithaAbstractiveMultidocumentSummarization2017,liImprovingMultidocumentsSummarization2014}.
This task has developed across many areas in computational linguistics for more
than six decades~\cite{luhnAutomaticCreationLiterature1958}.  Recently \llms\
have been influential and shown to achieve \acl{sota} performance in
summarization.  However, these models cannot condense a large number of long
documents given memory constraints.  Hallucination (erroneous and nonfactual
generated text) presents additional challenges with \llm\
summarizations~\cite{yehudaInterrogateLLMZeroResourceHallucination2024}.

In this work our goal is to automatically generate a \ds\ using clinical notes
after a patient leaves the hospital.  A \ds\ is a medical document that
explains a patient's illness, reason for their hospital stay, and treatment.
Physicians and clinicians hand-write exhaustive documentation during
hospitalization, which can be exploited to aid in authoring time-consuming
documentation.  \zfref{note-flow} illustrates this process by which previously
written clinical notes (note antecedents) are stored in the \ehr\ system and
later utilized to generate the \ds.

Clinical summarizations must be both faithful and traceable.  Generating \dss\
with these requirements highlights the difficulty of the task for admissions of
extended stay patients, which have \ac{ehr} notes that number in the hundreds.
Contemporary \ac{sota} methods, such as fine-tuning \acp{llm}, render the task
nearly impossible with the volume of information for patients with prolonged
hospitalizations.  In some cases, \acp{llm} might feasibly summarize clinical
documentation on a per note basis.  However, total textual content across all
handwritten notes of admissions easily extends past the limit of \acp{llm}'
context window.
Even the impressively large 2-million-token window of Gemini
2.0~\cite{anilGeminiFamilyHighly2024} is not large enough for lengthy
admissions.\footnote{The largest admission includes 1,233 notes in the
  \mimicname\ corpus.}.

Because of these memory constraints, other solutions are needed for the large
multi-document automatic summarization task of generating the \ds.  Most
abstractive methods inherently have a tendency to create less faithful
summaries because they typically formulate text as a probability distribution
over the vocabulary.  They also provide no traceable means of cross-referencing
the text of summarized documents.  The chosen method for our work is
extractive since it is both faithful and traceable, and thus, acceptable for
the clinical domain.

While previous methods have shown success at summarizing a single
section~\cite{adamsWhatSummaryLaying2021}, to our knowledge, there is no
peer-reviewed work that attempts to generate a complete \ac{ds} using \ac{ehr}
notes.  This motivates the extractive methods formulated in this work and
provides a baseline for future abstractive summarization.

The contributions of this work include
\begin{zlenumerateinline}
\item a faithful and traceable method to generate discharge summaries
\item the reusable \calsumurl[source code] to reproduce our results
\item the \mimic\ generated discharge notes with physician informal
  evaluations
\end{zlenumerateinline}

\graphStatMergeTab[t!]{\tabsize}
\zssec[related]{Related Work}

Summarization is a well-established area in
\ac{nlp}~\cite{zhangOptimizingFactualCorrectness2020,\liuct,\liaoct,saltonAutomaticAnalysisTheme1994,luhnAutomaticCreationLiterature1958}.

\zssubsec[rel-work:medical:ds]{Clinical Note Summarization}

To our knowledge, no other work exists that uses graph methods for summarizing
discharge notes.  However, the literature is rich with examples of clinical
note summarization that include both
longitudinal~\cite{hirschHARVESTLongitudinalPatient2015}, and
non-longitudinal~\cite{pivovarovAutomatedMethodsSummarization2015} note types,
two examples of mutual discipline interest.  Furthermore, the shared
understanding, agreement, and acknowledgment that faithful summarization is
necessary, but lacking, has been thoroughly
reviewed~\cite{zhangOptimizingFactualCorrectness2020}.

\citet{adamsWhatSummaryLaying2021} showed promising results in summarizing the
\textit{Brief Hospital Course} section.  However, for the single-section case,
the summarization of physician notes is perhaps the most interesting comparison
and potentially most impactful~\cite{gaoSummarizingPatientsProblems2022}.
Clinical notes were summarized by fine-tuning the
T5~\cite{raffelExploringLimitsTransfer2020} and
BART~\cite{lewisBARTDenoisingSequencetoSequence2020} \acl{sota} seq2seq models
and evaluated using the \bertscore~\cite{\bertscorect} and
\rouge~\cite{\rougect} scoring methods.

\zssubsec[rel-work:gen:amr]{Abstract Meaning Representation}

Interest in \amr\ has recently spanned across many
tasks~\cite{\liuct,bonialDialogueAMRAbstractMeaning2020,naseemDocAMRMultiSentenceAMR2022}.
\Ac{amr} graphs were later enriched with \propbank\ frames, which greatly
enhanced their expressiveness~\cite{palmerPropositionBankAnnotated2005}. Recent
achievements that use \ac{amr} models as the primary data representation
include work in \ac{nlg}~\cite{manningHumanEvaluationAMRtoEnglish2020},
\acl{mt}~\cite{blloshmiXLAMREnablingCrossLingual2020},
\acl{qa}~\cite{limKnowWhatYou2020}, and building logical
forms~\cite{galitskySummarizedLogicalForms2020}.

The well-known work of \citet{\liuct} used reduction methods with \ac{amr}
graphs for summarization.  In this work, the authors created a fully connected
graph that was used heuristically generate abstractive text.  This was later
broadened with a more comprehensive and robust \ac{amr} graph based realization
algorithm for multi-document
summarization~\cite{ogormanAMRSentenceMultisentence2018,\liaoct}.

This research is inspired by the work of \citet{\liuct} and \citet{\liaoct}
with regard to \ac{amr} graph reduction methods.  Our method differs in that it
leverages \calamr~\cite{\plcalamrct} to induce a graph by modeling it as flow
network~\cite{gaoFullyDynamicElectrical2022}, whereas their work builds on the
graph reduction methods of \citet{thadaniSentenceCompressionJoint2013}.  For
sentence comprehension, which re-frames the concept of commodity
flow~\cite{magnantiOptimalTrees1995} as indicator flow constraints for edge
inclusion.  For summarization, we leverage the \calamr\ method as it provides
tracability through \ac{amr} graph alignments.

\zssec[smy:ds]{Datasets}

The \mimiccite\ corpus and the \uicds\ were used for all experimentation using
the summarization methods described.  The \uicds\ is an IRB-approved private
dataset of 11,001 admissions and 607,872 notes, which include daily progress,
radiology, ECG and a variety of other notes from the \acl{uihealth} hospital.
Of the \mimic\ sample of
11,957 admissions, 113 admissions were processed.
\paulnote{I moved the issues with process to the Limitations section, but I
  don't like how early on we're discussing the ``lowest note count'' either.}

A total of 3,520 of the 11,957 \mimic\ admissions were
aligned~\zsseesec{smy:method} to create the Source Section
Dataset~\zsseesec{smy:srcsecds}.  \ztRef{graphStatMergeTab} shows the average
number of alignments across note antecedent and \ds\ components and the average
number of reentrancies per admission.  The ``alignable'' statistics are nodes
that are alignment candidates, such as concept and attribute notes.  The
``aligned'' statistics are those nodes with alignment edges.

Aligning the \uicds\ resulted in additional challenges.  The dataset has more
notes across category types compared to \mimic\ because the latter has only
\ac{icu} notes available~\cite{\pldsprovct}.
\paulnote{Because this was the summarization chapter, there are a lot of self
  references, but in cases like this, I don't know what else to do since I
  didn't get this fact from the literature, I got it from the physicians.}
The consequence of this more robust note variety is
that admission note counts are much higher, and therefore, take much longer to
align.  There is also a higher risk of missed alignments due to a potentially
higher rate of reentrancies (more than one in a path from the reentrancy to the
root), which lead to flow issues~\cite{\plcalamrct}.  Even though the \mimic\
alignments far outnumber the \uicds, the \uicds\ has many more reentrancies.

\zssec[smy:method]{Methods}

\calamrurl\ (\acl{calamr}) was leveraged to find clinical notes and candidate
sentences to use for summarization.  We refer the reader to the paper by
\citet{\plcalamrct}, but we give a brief overview here.  The \calamr\ method
first parses the source text into a single connected graph of \ac{amr} sentence
graphs.  It then does the same for the summary text.  These two graphs start as
separate components that become one bipartite graph.

\zfadd[t!][\columnwidth]{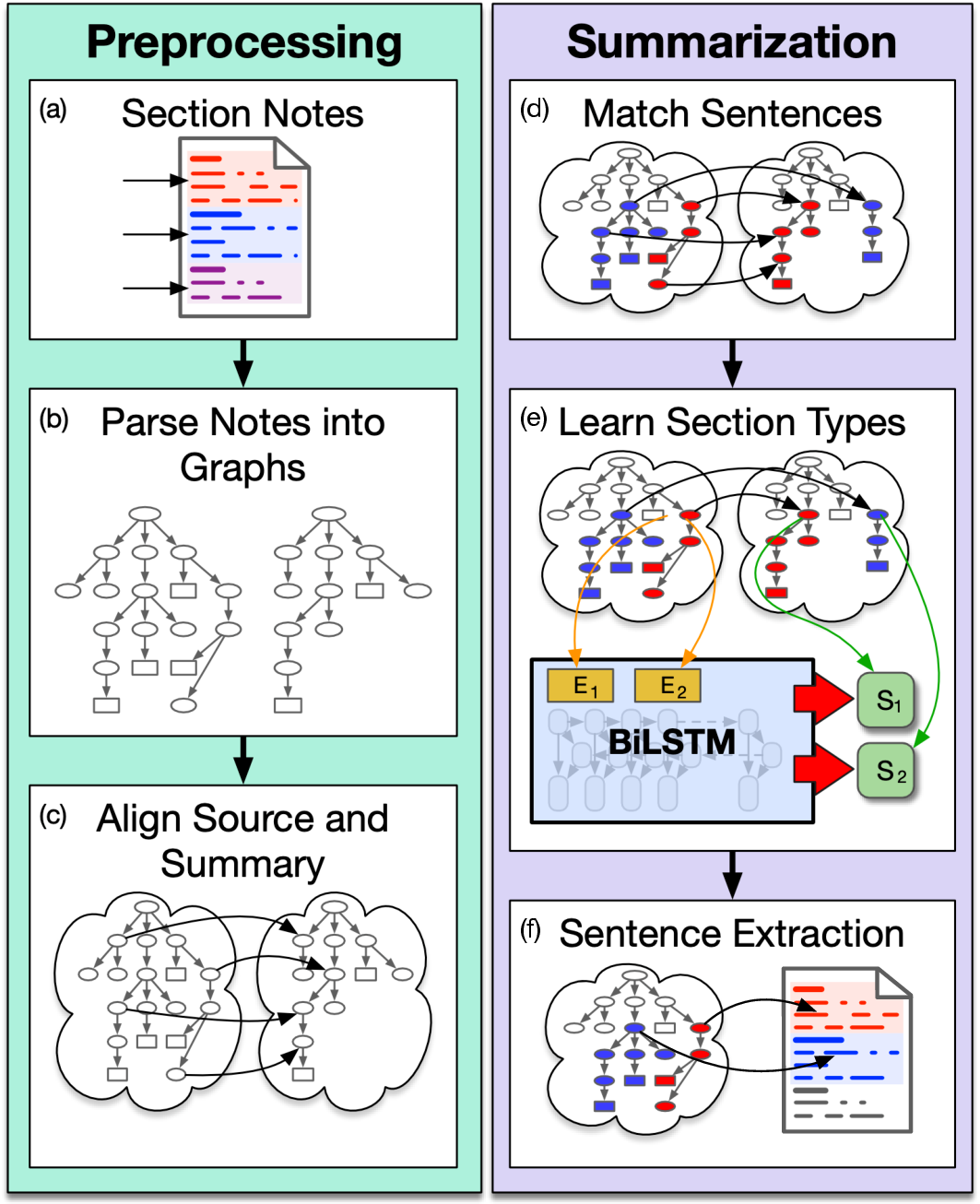}[Pipeline Overview]{Clinical notes are
  first preprocessed (left) into learning examples for a summarization model
  (right).}

Nodes are connected, as bipartite edges, if their semantic similarity's
neighborhood exceeds a threshold.  This similarity measure is calculated based
on embeddings assigned to concept and attribute \ac{amr} nodes and
\propbank~\cite{\propbankct} roles and role set edges.  The similarity measures
are also used as the information gain across the connected graph and all
subgraphs of each in the max flow
algorithm~\cite{gaoFullyDynamicElectrical2022,ford1962flows}.  The assigned
flow values to each bipartite edge lead to the ``starvation'' of low
information subgraphs.  Subgraphs are effectively removed by setting low flow
alignment edge capacities to zero.

\zfaddtc[][\textwidth]{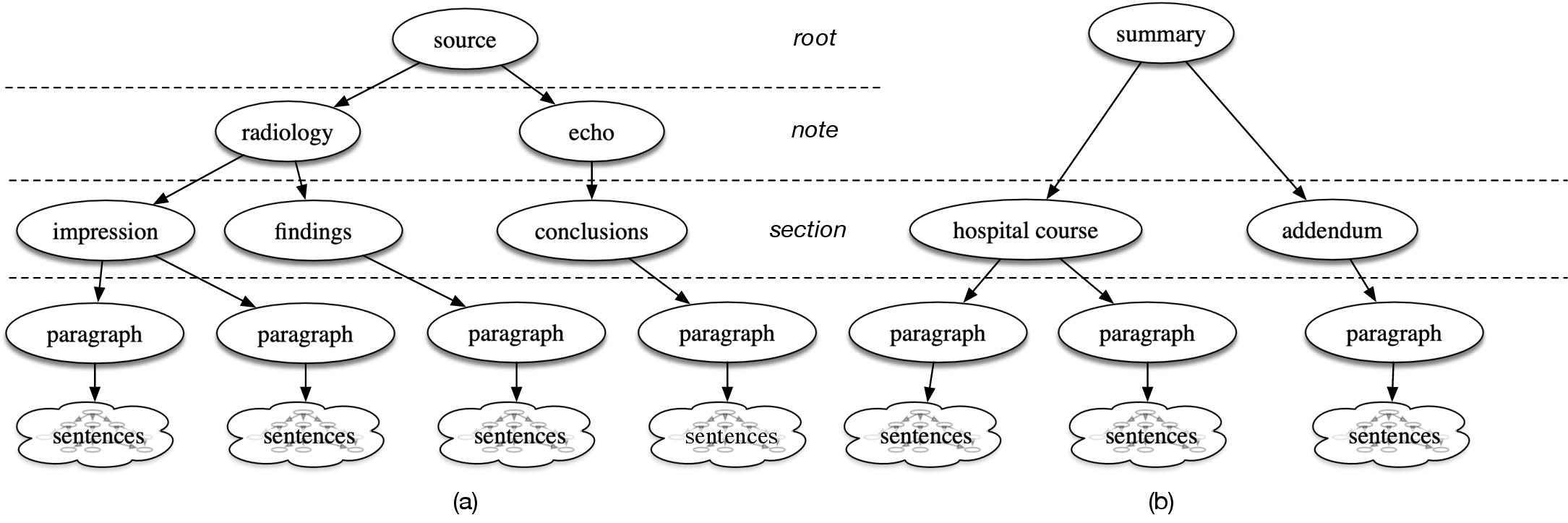}[Admission Graph]{An admission graph of
  the note antecedents (a), and the \ds\ (b).}[eps]

\zfaddtc[b!][\textwidth]{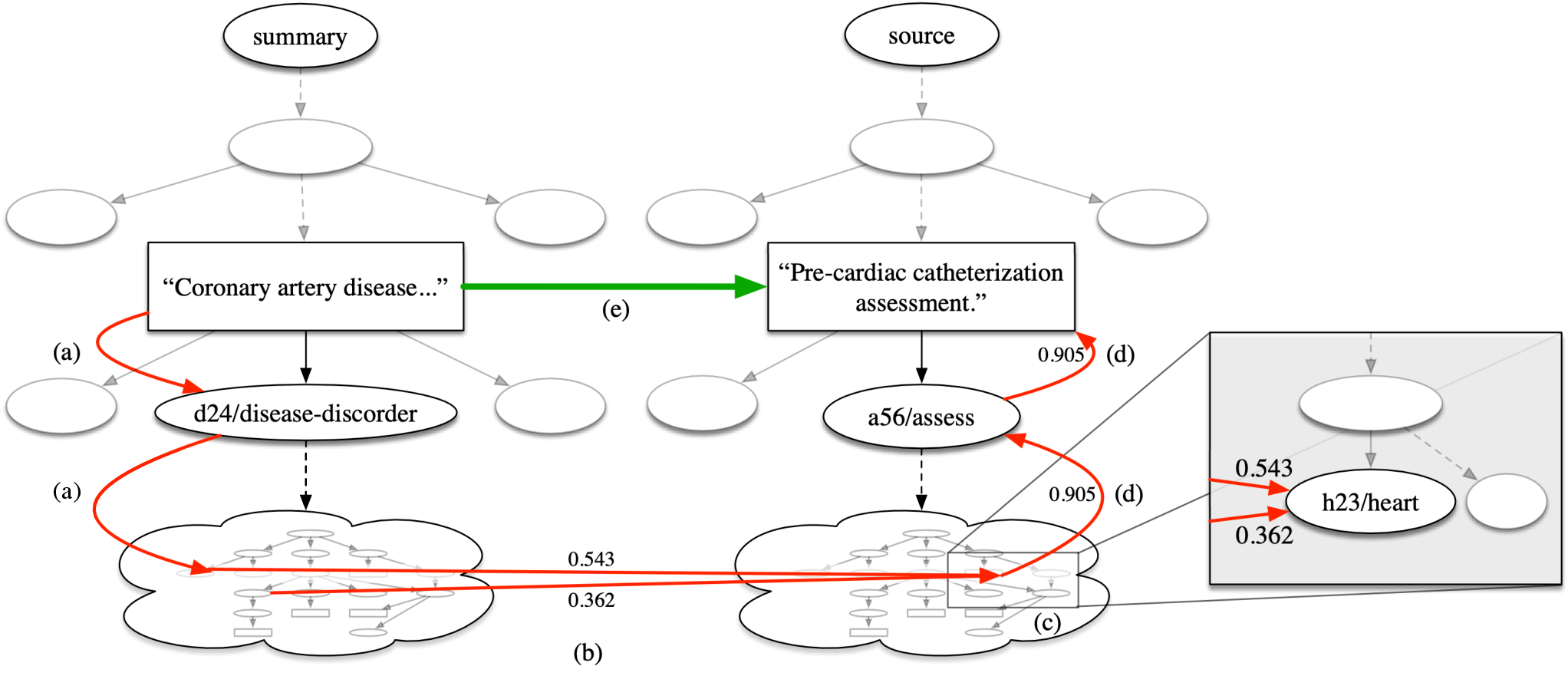}[Sentence Matching Algorithm]{%
  The path ({\color{red}red}) of alignment flow from the source to the summary
  for sentence. The enlarged box shows two incoming alignment flows from the
  source into the \texttt{heart} concept with a combined flow of 0.905. The
  green arrow represents a match candidate as a result of this alignment flow
  and the path to their respective sentences.}

\calamr\ was used to create supervised training examples using the flow data of
the alignment graphs to match note antecedent source sentences to \ds\
sentences.  A model we refer to as the Source Section Model classified the \ds\
section type of notes written prior to the \ds\ (note antecedent) to the source
sentence~\zsseesec{smy:method:model}.  The source section model then used the
matches to learn what to add to the summary.  Each note antecedent source
sentence was then assigned to the section of the matched \ds\ sentence.  This
was then used as the label in a classification \ac{nn} model.

An overview of the pipeline follows \zfsee{pipeline-overview}:
\begin{zlpackedenum}
\item Preprocess notes to generate a dataset for supervised
  training.\zsstepadd{preproc}
  \begin{zlpackeditemize}
  \item[(a)] Construct an admission graph from a subset of \mimiccite\
    admissions~\pipe{a}.
  \item[(b)] Create \amr\ sentences using a text-to-graph parser.
  \item[(c)] Use \calamr\ to create an alignment graph for each admission.  This
    includes the note antecedents connected to the \ds~\pipe{c}.
  \end{zlpackeditemize}

\item Train a summarization model using the dataset created in
  \zsstepref{preproc}, then use it to summarize.
  \begin{zlpackeditemize}
  \item[(d)] Label note antecedent sentences with \ds\ section types using the
    alignments~\pipe{d}.\zsstepadd{smy:method:matchcorp}
  \item[(e)] Train a supervised sentence classification model using the labels
    created
    in~\zsstepref{smy:method:matchcorp}~\pipe{e}.
  \item[(f)] Use the trained model to add note antecedent sentences by \ds\
    section to the generated note.
  \end{zlpackeditemize}
\end{zlpackedenum}

\zssubsec[smy:method:admgraph]{Admission Graph}

A patient is admitted to the hospital upon entering for any administered
healthcare services.  From the healthcare perspective, this admission includes
what is done to the patient for the duration of the hospital stay.  The
admission graph is a semantic representation of a patient's hospital stay.  It
is composed of two disconnected graph components: all the antecedent notes for
the admission and the \ds.

The sentences of all antecedent notes are parsed into \amr\ graphs, and then
connected to create the source graph.  Likewise, the \ds\ is parsed into \amr\
graphs, and when connected, become the summary graph.  These two disconnected
components follow the structure of the source and summary components of the
bipartite graph described by \citet{\plcalamrct}.  However, document nodes that
represent note categories, note sections and clinical text paragraphs are used
between the roots and their respective \amr\ subgraphs as shown in
\zfref{adm-graph}.  The note antecedent source has a note category level,
whereas the summary component's root represents the \ds.

The \spacyurl\ and \scispacyurl\ libraries was used to tokenize, sentence chunk
and tag biomedical and non-biomedical named entities.
MedCAT~\cite{kraljevicMultidomainClinicalNatural2021} was used to link token
spans to \umls\ \acfp{cui} that aid in graph aligning their text-to-graph concept
nodes.

Previous methods have used concept merging to join \ac{amr}
sentences~\cite{leeAnalysisDocumentGraph2021}.  However, we joined the \ac{amr}
parser's selected sentence roots to their corresponding paragraph nodes and
used Coreference Resolution in place of concept node merging to avoid loss of
data.

\zssubsec[smy:sentmatch]{Sentence Matching Algorithm}

The sentence matching algorithm uses the \calamr\ alignments to identify the
sentences that best represent the summary.  This classification is based on the
sentence-to-sentence information gain from the aligned graph flow network.
\zfref{adm-match} shows how the source sentence, \textit{Pre-cardiac
  catheterization assessment.}  matches with the \ds\ sentence
\textit{``Coronary artery disease, status post coronary artery bypass
  grafting,''} by creating paths through the graph from a source sentence to a
summary sentence.  Each sentence connected in this way becomes a candidate.

The sentence matching algorithm follows:
\begin{zlpackedenum}
\item For each \ds\ sentence in the reduced graph, use a depth-first search to
  index aligned nodes (\zfrefsub{adm-match}{a}).\label{step:smy:sentmatch:dfs}
\item For each indexed node in step~\ref{step:smy:sentmatch:dfs}, traverse the
  alignment edge to source nodes in the note antecedent component
  (\zfrefsub{adm-match}{b}).\label{step:smy:sentmatch:align}
\item Annotate aligned source nodes indexed in
  step~\ref{step:smy:sentmatch:align} with alignment flows from \ds\
  component edges (\zfrefsub{adm-match}{c}).
\item Associate the aligned node summary annotations for each respective
  sentence in the source component (\zfrefsub{adm-match}{d}).
\item Create a sentence match candidate between the source and summary
  sentences (\zfrefsub{adm-match}{e}).
\item Sort the source sentences by the sum of the flow from each summary
  sentence.\label{step:smy:sentmatch:srt}
\item Match sentences based on the flow from each summary to source
  sentence.\label{step:smy:sentmatch:match}
\item All remaining unmatched note antecedent sentences are given the
  \texttt{no-section} label.
\end{zlpackedenum}

\begin{table*}[t!]
  \centering
  \tabsize%
  \begin{tabulary}{\textwidth}{l@{\hskip 0.9cm} r r r@{\hskip 0.9cm} r r r}
    \toprule
    & \multicolumn{3}{c}{\textbf{\mimicname}} & \multicolumn{3}{c}{\textbf{\uihealthname}} \\
    \cmidrule{2-7}
    \textbf{Section}           & \textbf{Train}   & \textbf{Test}   & \textbf{Validation}   & \textbf{Train}   & \textbf{Test}   & \textbf{Validation}   \\
    \hline
    \sectionLabelCountsMergedTab%
    \bottomrule
  \end{tabulary}
  \caption[\sectionLabelCountsMergedTabHead]{\textbf{\sectionLabelCountsMergedTabHead{}.} \sectionLabelCountsMergedTabCaption}
  \label{tab:sectionLabelCountsMergedTab}
\end{table*}

Once the source sentences are paired with distributions of summary sentences by
flow in step~\ref{step:smy:sentmatch:srt} each source sentence is matched with
zero or more summary sentences.  In step~\ref{step:smy:sentmatch:match}, a
source sentence is matched with the summary sentence that has the maximum flow
determined by the minimum sent flow hyperparameter.  The matched summary
sentence is then eliminated as a candidate for matching with any other source
sentence.  Finally, the source sentences are tagged with the section of the
matched summary sentence.  Upon completion, antecedent sentences are tagged
with the \ds\ section to which it should be added.  For example, a radiology
antecedent note may a sentence marked a \textit{Brief Hospital Course} \ds\
section type. This sentence will then be added to that section during the
generation process.


\zssubsec[smy:srcsecds]{Source Section Dataset}

\zfadd[b!][\columnwidth]{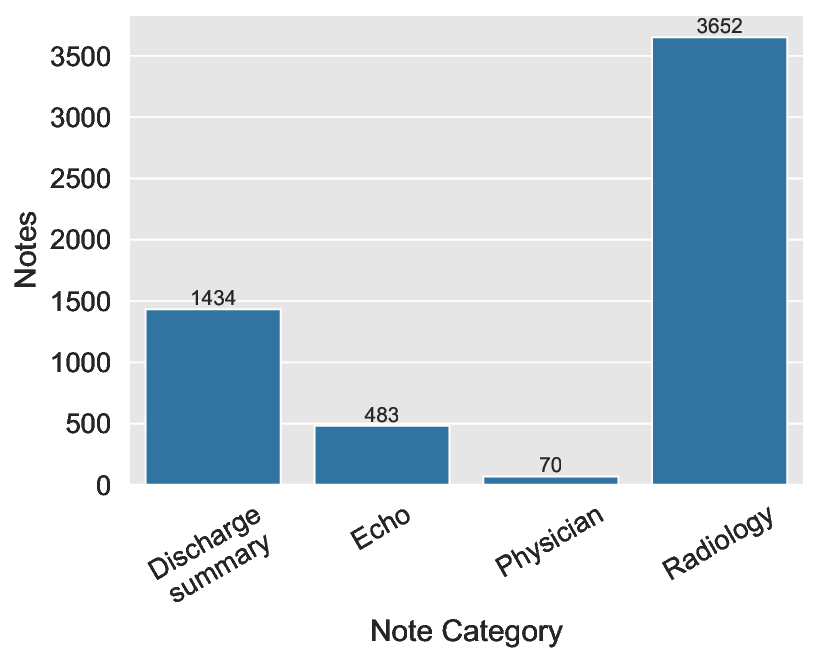}[MIMIC Matched Sentence Notes]{%
  Counts of notes per admission in the Source Section Dataset.}[eps]

We refer to the set of notes that were successfully aligned (described
in~\zssecref{smy:ds}) as the Source Section Dataset.  The sentence matching
algorithm described in \zssecref{smy:sentmatch} was used to automatically pair
sentences from note antecedents to \dss\ of this dataset.  Each sentence pair
of each admission graph will be used to train an extractive summary model
\zsseesec{smy:method:model}.  The note counts by categories are given in
\zfref{noteCount}.

The selected \ds\ sections were based on what were considered most necessary
and beneficial for summarization by a physician authoring the note, by a
clinical informatics fellow and a 4\textsuperscript{th} year medical student.
The physician-selected \ds\ sections and their counts are given in
\ztref{sectionLabelCountsMergedTab}.  Most notable is the imbalance between the
section labels and \texttt{no-section} label.  This high disparity leads to a
terse generated \ds, which is explained further in \zssecref{smy:res}.  The
same process was used when selecting sections in the \uicds.
\zsapxref{cont:mimic} and \zsapxref{cont:uihealth} give the matched sentence
candidate contingency tables\footnote{The contingency tables based source
  sentences with their associated distribution (in terms of alignment graph
  flow) of summary sentences.}.


\zssubsec[smy:method:gen]{Discharge Summary Generation}

The \dss\ were generated using the source section
model~\zsseesec{smy:method:model} trained on the Source Section Dataset.  The
note antecedents\ of the Source Section Dataset's test set were used as input
to the source section model.  Sentences were added to the predicted section in
the generated \ds\ or discarded if the \texttt{no-section} label was predicted.

The \uicds\ was used as a development set by tuning the \calamr\
$k$\textsuperscript{th} order neighbor set hyperparameter
(\ensuremath{\mathbf{\Lambda}}) to include more network neighborhood semantic
information.  The minimum sentence flow hyperparameter
(\ensuremath{\mathbf{\mu_{s}}}) was also adjusted to increase the output to 248
aligned admissions with higher quality.

The \mimic\ trained summarization model yielded 133 automatically generated
\dss\ and the \uicds\ model generated five.  The alignment challenges described
in \zssecref{smy:ds}, such as missing \dss\ and GPU memory constraints, show
the difficulty of hospitalization summarization.  Further discussion of these
challenges are described in \zssecref{smy:res}.


\zssubsubsec[smy:method:model]{Source Section Model}

Once the sentence matching algorithm was used to assign labels to source
sentences \zsseesec{smy:sentmatch} a \ac{bilstm} was trained to learn the \ds\
section type of each note antecedent source sentence.  A section, such as
\textit{Hospital Course}, is a label predicted by the model indicating that not
only should the sentence be added, but to which section in the \ds\ to add it.
A label of \texttt{no-section} means the sentence is to be discarded.

A \ac{bilstm}~\cite{\bilstmct} was used for learning the sentence section
classification.  The \gatortron~\cite{\gatortronct} clinical embeddings, the
note antecedent's note category, and the section type were used as input
features to the model.  Because of the data input size~\zsseesec{smy:ds} the
model's static embeddings were used in place of fine-tuning.  A fully connected
linear layer was added between the \ac{bilstm} and the output layer.  The
\ac{bilstm} layer had a hidden size of 500 parameters, a dropout of $p = 0.15$,
a learning rate of $5 \times 10^4$ and used gradient clipping.  The model was
set to train for 30 epochs and converged at 24 epochs.

\zssec[smy:method:eval]{Experimental Setup}

The standard set of quantitative \ac{ml} performance metrics were used to
evaluate the source section model.  Automatically metrics such as
\rougename~\cite{\rougect}, \acs{bleu}~\cite{\bleuct} and
\acs{bertscore}~\cite{\bertscorect}, are of little help with such a large set
of disjoint text less than half of the \ehr\ notes is represented in the
\ds~\cite{adamsWhatSummaryLaying2021,\pldsprovct}.

\zssubsec{Limitations of Automatic Evaluation Metrics}

To demonstrate the limited effectiveness of computed automated evaluation
metrics, we compared the \ehr\ records with the discharge summary\footnote{The
  note antecedents (sorted by chart date) were concatenated for comparison.}.

\begin{table}[h!]
  \centering
  \tabsize%
  \begin{tabulary}{\textwidth}{l@{\hskip 1.5cm} l}
    \toprule
    \textbf{Metric} & \textbf{Value} \\
    \hline
    \autoEvalMetricsTab%
    \bottomrule
  \end{tabulary}
  \caption[\autoEvalMetricsTabHead]{\textbf{\autoEvalMetricsTabHead{}.} \autoEvalMetricsTabCaption}
  \label{tab:autoEvalMetricsTab}
\end{table}

\ztRef{autoEvalMetricsTab} shows very low scores \rougename\ and \acs{bleu}
across the original (unmodified) \mimicname\ antecedents with the discharge
summaries.  For this reason we believe human evaluation is appropriate for
judging the effectiveness of generated documentation given the depth,
complexity and technical jargon found in clinical notes.

We used human quantitative and qualitative evaluation for a more reliable
metric to better understand the effectiveness of the generated notes.  This
evaluation of \dss\ on the Source Section Dataset's test set was evaluated by a
clinical informatics fellow and a 4\textsuperscript{th} year medical student.
Each generated \ds\ was ranked using a Likert
scale~\cite{likertTechniqueMeasurementAttitudes1932} as an integer value
ranking in the range 1 to 5 with five as the highest.
\ztRef{informalevalquestab} lists the questions asked for the informal
evaluation.

\informalevalquestab{\tabsize}

\zssec[smy:res]{Results}

The source section model results are summarized in \ztref{dsGenSummaryTab}.
The weighted F1 score of 88.72 on the \mimic\ trained corpus shows good results
for the sentences' \ds\ section classification.  However, we see a low macro F1
of 20.41.  Another reason for a high weighted and micro F1 but low macro F1 is
that the majority label, the \texttt{no-section} label, dominated as shown
in~\zfref{noteCount}.

\dsGenSummaryTab[t]{\tabsize}

The model trained on the \uicds\ shows lower results.  This might be partly due
to the smaller dataset or the higher rate of reentrancies as shown
in~\ztref{graphStatMergeTab} and discussed in~\zssecref{smy:ds}.  The fact that
\mimic\ is a curated dataset is the most likely reason the results are higher
compared to \uicds, which is unmodified and contains \ac{hphi}.

The informal evaluation of the 133 generated \dss\ trained on the \mimic\
corpus \ztsee{informalEvalTab}.  The evaluation illuminates the difficulty of
the task. Despite low scores on sectioning, completeness and preference the
generated summaries provide a readability of just over 3 and a perfect
correctness score (5), which is what we expect from a faithful summary.

\informalEvalTab[h]{\tabsize}

\zssubsec[smy:res:disc]{Discussion}

The label imbalance in the Source Section Dataset might be attributed to the
sparsity of \calamr's alignments.  If this were the case, we could adjust the
hyperparameters\ of \calamr\ to produce more sentence matches.  However, the
lack of alignment could be justified by the lack of notes (other than those
from the \ac{icu} department) present in the \mimic\ corpus.  The misalignment
could also be attributed in cases where the physician writes from personal
experience with the patient that is otherwise lacking from the \ehr\ notes.

The five \dss\ produced by the model trained on the
\uicds~\ztsee{informalEvalTab} show better completeness but slightly lower
readability.  A higher sectioning score was given to the \uicds\ despite the
fact that the \medsecid\ model was trained on \mimic.  This implies the
\medsecid\ is able to section the \uicds\ notes or the source section model is
able to predict sections based on other factors such as better alignments.  A
de-identified automatically generated \ds\ \zfref{ds-gen-sym-uic} and its gold
counter-part are given in \zsapxref{ds-gold-sym-uic} and
\zsapxref{ds-gen-sym-uic}.

\zssec[smy:conc]{Conclusion}

We have presented a new extractive method of multi-section automatic
summarization; a significant step forward in generating faithful and traceable
summaries for clinical documentation.  The source section model shows promising
results.  The generated training data from the \calamr\ alignments produces a
model that learns how to classify for extractive summarization.  This is
further qualified by the informal evaluation with high correctness scores and
reasonable readability scores.
Shortcomings of the data used to train the models led to challenges that
affected performance and brought to light certain limitations.  In the case of
the \mimic\ data, the issue of missing notes led to worse summarizations from
missing hospitalization notes.
The summaries evaluated are faithful in that only content from the source text
is added to the summary.  They are traceable in how each sentence can be traced
back via the \calamr\ alignments.  Recommendations with respect to recent
\ac{sota} breakthroughs also inspire hope to bridging the gap of the difficulty
of the summarization task.



\zavenenddoc%

\clearpage
\onecolumn
\zavenappendix%
\setlength\intextsep{0pt}

\zsapx[ds-gen-sym-uic]{Generated Discharge Summary}
\vspace{\apxheadspace}
\begin{zzdsmultipage}[\textwidth]{ds-gen-sym-uic}{Generated Discharge Summary}{%
    A \uihealthname\ de-identified automatically generated discharge summary.}
History of present illness:
9:36 AM Status: Attested Editor: [**Doctor First Name**] [**Doctor Last Name**], MD (Resident) Related Notes: Original Note by [**Doctor First Name**] [**Doctor Last Name**], MD (Resident) filed at [**Date**] 4:01 PM Cosigner: [**Doctor First Name**] [**Doctor Last Name**], MD at [**Date**] 12:46 PM Consult Orders 1. Inpatient consult to General Neurology [[**Correspondence ID**]] ordered by [**Doctor Last Name**] [**Doctor Last Name**], MD at [**Date**] 0754 Attestation signed by [**Doctor First Name**] [**Doctor Last Name**], MD at [**Date**] 12:46 PM Stroke Attending: Pt eloped prior to being seen. Blurred Vision and Extremity Weakness [**First Name**] [**Last Name**] is a 47 y.o. male with PMH HTN, HLD, previous stroke [**Year**] (R ACA-MCA watershed), presenting to ED with L sided weakness, LLE numbness, and blurry vision b/ l. Patient reports symptoms started acutely on Friday [**Date**] around 4 pm while driving causing him to have to pull over. He decided to try to sleep it off. After waking up the following morning with no improvement, he went to [**Hospital Name**] ED where he was seen by neurology, but left AMA as he felt he was being asked the same questions repeatedly and nothing was getting done. CTH at [**Hospital Name**] was without ICH, reportedly showed wedge shaped hypodensity in frontal lobe likely from chronic infarct. The patient reports that the left sided weakness has improved somewhat today, but he still endorses b/ l blurry vision with constant white floaters, as well as numbness/ tingling in his LLE. CTA Hwith multifocal narrowing of b/ l ACAs as well as R M1 focal narrowing. MRI B w/o showing acute ischemia in medial R temporal lobe involving the posterior limb of R internal capsule as well as possibly the thalamus, and redemonstrating old R frontal ACA-MCA watershed infarct and old L occipital cortical infarct. Patient to be admitted to stepdown under stroke service. Patient with poorly controlled HTN and HLD, not taking any medications for the past 5 months as he says he had trouble getting primary care appointment for prescription renewals.

Physical examination:
NIHSS is 2 (LUQ quadrantopia, LLE numbness). SBP gets up to 200s per patient. Labs today significant for Troponin of 0.77. EKG showing T inversions in V5 and V6. Cardiology consulted. Patient denying CP at this time. Prior stroke/ TIAs (date, description): R ACA-MCA waterhsed stroke in [**Year**] per ED, worked up at [**Hospital Name**], on DAPT Vascular risk factors: HTN, HLD, prior stroke Past Medical/ Surgical history: Past Medical History: Diagnosis Date Hypertension
\end{zzdsmultipage}

\vspace{.5cm}

\zsapx[ds-gold-sym-uic]{Gold Discharge Summary}
\vspace{\apxheadspace}
\begin{zzdsmultipage}[\textwidth]{ds-gold-sym-uic}{Gold Discharge Summary}{%
    The physician hand written gold \uihealthname\ de-identified discharge %
    summary.  White space that spanned longer than three lines was reduced to two.}
Discharge Summary by [**Doctor First Name**] [**Doctor Last Name**], MD at [**Date**] 6:00 AM

Author: [**Doctor First Name**] [**Doctor Last Name**], MD Service: Neuro Critical Care Author Type: Resident
Filed: [**Date**]  6:09 PM Date of Service: [**Date**]  6:00 AM Status: Attested
Editor: [**Doctor First Name**] [**Doctor Last Name**], MD (Resident) Cosigner: [**Doctor First Name**] [**Doctor Last Name**], MD at [**Date**] 12:45 PM
Attestation signed by [**Doctor First Name**] [**Doctor Last Name**], MD at [**Date**] 12:45 PM  Stroke attending: I have reviewed the above discharge summary and agree with the assessment.

Pt eloped before I could staff the pt.

[**Hospital Name**]
Discharge Summary

Patient: [**First Name**] [**Last Name**]
Admission Date: [**Date**]
Discharge Date: [**Date**]
Discharge Disposition: Left Against Medical Advice
Discharge Service: Stroke
Discharge Attending: [**Doctor First Name**] [**Doctor Last Name**], MD
Primary Diagnosis: Acute R medial temporal and internal capsule/thalamic stroke

Other Active Diagnoses
Diagnosis Date Noted POA
- Troponin level elevated [**Date**] Yes
Priority: High
- Stroke (CMS/HCC) [**Date**] Yes

Hospital Course
HPI:[**First Name**] [**Last Name**] 47 y.o.PMH HTN, HLD, previous stroke [**Date**] (R-MCA watershed), presenting to ED with L sided weakness, LLE numbness, and blurry vision b/l. Patient reports symptoms started acutely on Friday [**Date**] around 4pm while driving causing him to have to pull over. He decided to try to sleep it off. After waking up the following morning with no improvement, he went to [**Hospital Name**] ED where he was seen by neurology, but left AMA as he felt he was being asked the same questions repeatedly and nothing was getting done. CTH at [**Hospital Name**] was without ICH, reportedly showed wedge shaped hypodensity in frontal lobe likely from chronic infarct. The patient reports that the left sided weakness has improved somewhat today, but he still endorses b/l blurry vision with constant white floaters, as well as numbness/tingling in his LLE. NIHSS is 2 (LUQ quadrantopia, LLE numbness). CTA Hmultifocal narrowing of b/l ACAs as well as R M1 focal narrowing.B w/oacute ischemia in medial R temporal lobe involving the posterior limb of R internal capsule as well as possibly the thalamus, and redemonstrating old R frontal ACA-MCA watershed infarct and old L occipital cortical infarct. Patient to be admitted to stepdown under stroke service.

Patient with poorly controlled HTN and HLD, not takingthe past 5 monthshe says he had trouble getting primary care appointmentprescription renewals. SBP gets up to 200s per patient. Labs today significant for Troponin of 0.77-0.62. EKG showing T inversions in V5 and V6. Cardiology consulted. Patient denying CP at this time. Cardio recommending trending EKG/trop until downtrend and ordering Echo.

Patient appeared to have left before he was evaluated [**Date**] AM.

Pertinent Physical Exam At Time of Discharge
Physical Exam
PATIENT NOT EXAMINED PRIOR TO DISCHARGE

Test Results Pending At Discharge

Discharge Medications

No medications have been prescribed.

Issues Requiring Follow-Up
- Patient not evaluated prior to discharge
Outpatient Follow-Up Appointments
No future appointments.
Referrals
No orders of the defined types were placed in this encounter.

Completed Consults:
Consults Ordered This Encounter
Procedures
- Inpatient consult to General Neurology
- Inpatient consult to Cardiology
\end{zzdsmultipage}

\vspace{.5cm}

\zsapx[cont:mimic]{\mimicname\ Contingency Table}
\vspace{\apxheadspace}
\zfaddtc[h!][4.5in,height=1.9in]{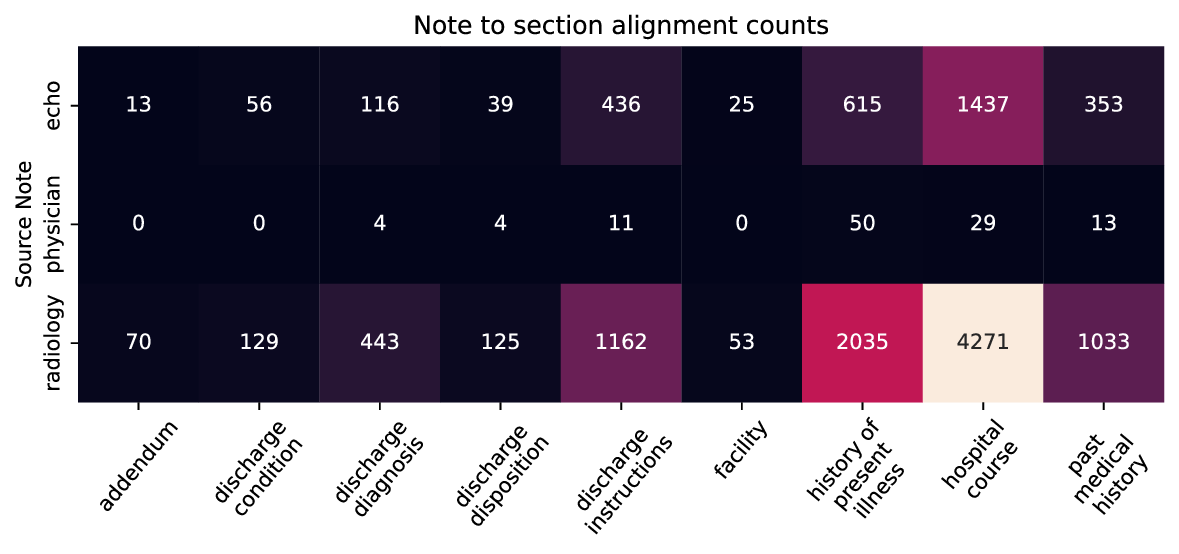}[\mimicname\ Note to Section
Alignment Contingency]{The contingency table shows \calamr\ alignment flow of
  data from the note antecedents to the sections of the \ds.}


\clearpage\newpage

\zsapx[cont:uihealth]{\uihealthname\ Contingency Table}
\vspace{\apxheadspace}
\begin{figure}[h!]
  \begin{center}
    \includegraphics[height=2.4in,angle=-90]{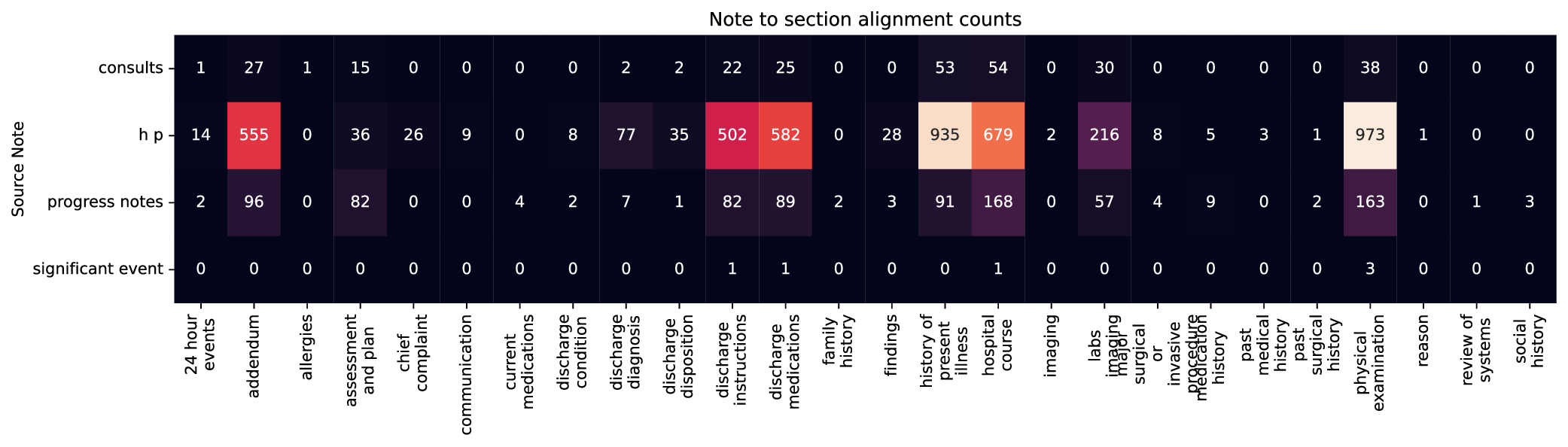}
  \end{center}
  \caption[\uihealthname\ Note to Section Alignment Contingency]%
  {\textbf{\uihealthname\ Note to Section Alignment Contingency.}  The
    contingency table shows \calamr\ alignment flow of data from the note
    antecedents to the sections of the \ds.}%
  \label{fig:uic-note-sec-contingency}
\end{figure}



\end{document}